\documentclass[10pt,twocolumn,letterpaper]{article}

\usepackage[pagenumbers]{iccv} 
\usepackage[most]{tcolorbox}
\usepackage{cuted}
\usepackage{float}
\usepackage{tabularx}
\usepackage{stfloats}

\usepackage[dvipsnames]{xcolor}
\usepackage{colortbl}

\newcommand{\methodname}[0]{\texttt{{APC}}}

\definecolor{iccvblue}{rgb}{0.21,0.49,0.74}
\usepackage[pagebackref,breaklinks,colorlinks,allcolors=iccvblue]{hyperref}
\usepackage{multirow}
\usepackage{multicol}
\renewcommand{\thefootnote}{\fnsymbol{footnote}}

\title{Perspective-Aware Reasoning in Vision-Language Models \\via Mental Imagery Simulation}

\author{
Phillip Y. Lee\textsuperscript{1} $\quad$
Jihyeon Je\textsuperscript{2} $\quad$
Chanho Park\textsuperscript{1} $\quad$
Mikaela Angelina Uy\textsuperscript{3} $\quad$\\
Leonidas Guibas\textsuperscript{2} $\quad$
Minhyuk Sung\textsuperscript{1} \\[0.2em]
\textsuperscript{1}KAIST $\quad$ \textsuperscript{2}Stanford University $\quad$ \textsuperscript{3}NVIDIA
}

\begin{document}

\twocolumn[{%
\renewcommand\twocolumn[1][]{#1}%
\maketitle
\begin{center}
    \centering
    \captionsetup{type=figure}
    \vspace{-1.5\baselineskip}
    \includegraphics[width=\textwidth]{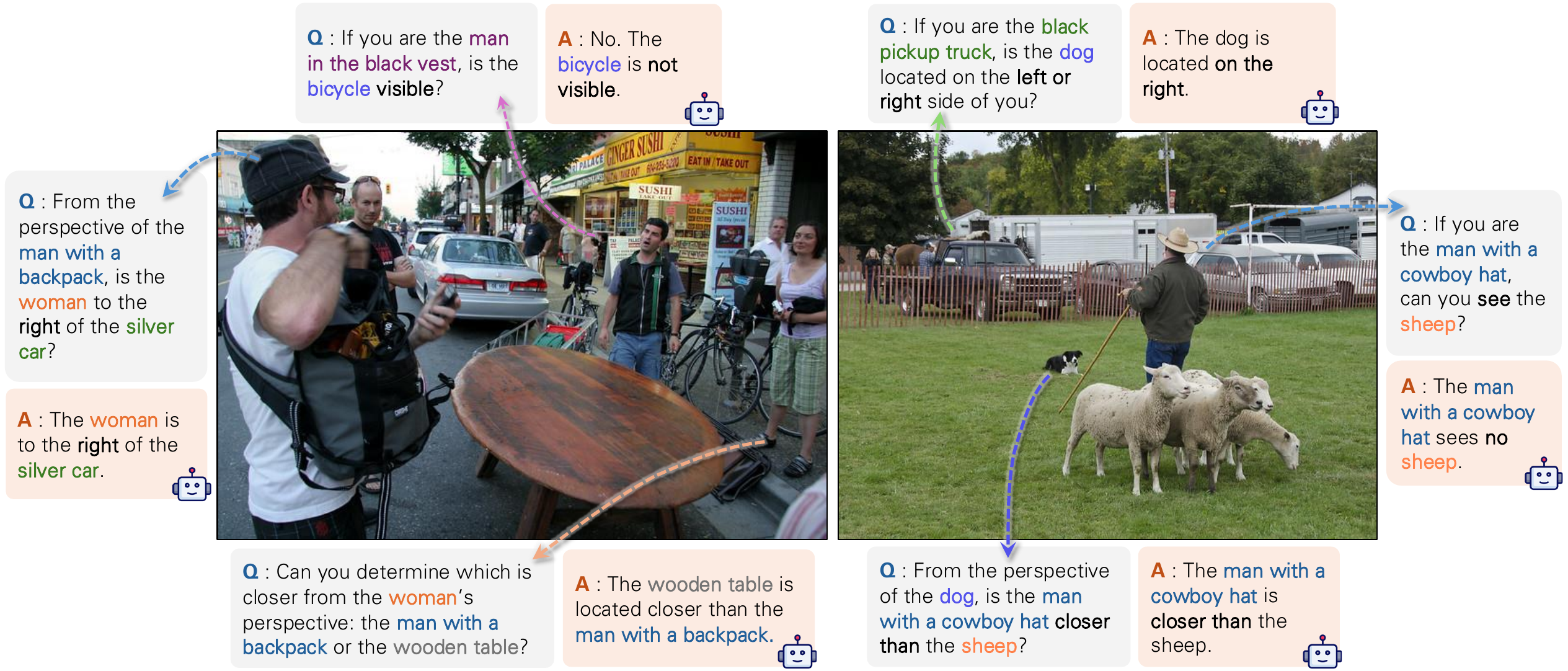}
    \vspace{-\baselineskip}
    \captionof{figure}{We introduce \textbf{Abstract Perspective Change (\methodname)}, a framework that empowers VLMs to adopt arbitrary perspectives for spatial reasoning. As demonstrated by the examples above, \methodname~significantly enhances VLM's ability to \emph{imagine a scene from alternative viewpoints}, overcoming the inherent egocentric bias that constrains the spatial reasoning of existing VLMs to the camera's viewpoint.}
    \label{fig:teaser}
\end{center}
}]

\def\thefootnote{1}\footnotetext{Correspondence: Phillip Y. Lee {\tt (phillip0701@kaist.ac.kr)} and Minhyuk Sung {\tt (mhsung@kaist.ac.kr)}}\def\thefootnote{\arabic{footnote}}

\maketitle

\begin{abstract}
We present a framework for perspective-aware reasoning in vision-language models (VLMs) through mental imagery simulation. Perspective-taking---the ability to perceive an environment or situation from an alternative viewpoint---is a key benchmark for human-level visual understanding, essential for environmental interaction and collaboration with autonomous agents. Despite advancements in spatial reasoning within VLMs, recent research has shown that modern VLMs significantly lack perspective-aware reasoning capabilities and exhibit a strong bias toward egocentric interpretations. To bridge the gap between VLMs and human perception, we focus on the role of mental imagery, where humans perceive the world through abstracted representations that facilitate perspective shifts. Motivated by this, we propose a framework for perspective-aware reasoning, named Abstract Perspective Change (\methodname), that effectively leverages vision foundation models, such as object detection, segmentation, and orientation estimation, to construct scene abstractions and enable perspective changes. Our experiments on synthetic and real-image benchmarks, compared with various VLMs, demonstrate significant improvements in perspective-aware reasoning with our framework, further outperforming fine-tuned spatial reasoning models and novel-view-synthesis-based approaches. Our project page is at \url{https://apc-vlm.github.io/}.
\end{abstract}    
\vspace{-\baselineskip}
\section{Introduction}
\label{sec:intro}
\vspace{-0.3\baselineskip}
Vision-language models (VLMs) have made remarkable progress, positioning themselves as a crucial backbone for general-purpose physical AI agents. The growing research efforts to improve VLMs’ spatial reasoning capabilities~\cite{tong2024cambrian1fullyopenvisioncentric, cheng2024spatialrgpt, chen2024spatialvlm, ma2024spatialpin} reflect this potential. Early VLMs were limited to basic tasks such as visual question answering (VQA) and image captioning~\cite{li2023blip, li2022blip, alayrac2022flamingo}. However, recent advancements have enabled them to perform complex visual reasoning~\cite{hurst2024gpt, team2024gemini, liu2023llava, Qwen2.5-VL, deitke2024molmo, hu2024visual} and extract spatial properties, including spatial relationships, relative sizes, and distances~\cite{li2024llavaonevision, tong2024cambrian1fullyopenvisioncentric}. Further techniques such as instruction-tuning and vision-centric adapters, have expanded their capabilities, allowing depth-aware~\cite{chen2024spatialvlm, cai2024spatialbot} and region-aware~\cite{cheng2024spatialrgpt, heo2025omni} spatial reasoning.

\begin{figure}[h!]
  \centering
  \includegraphics[width=\linewidth]{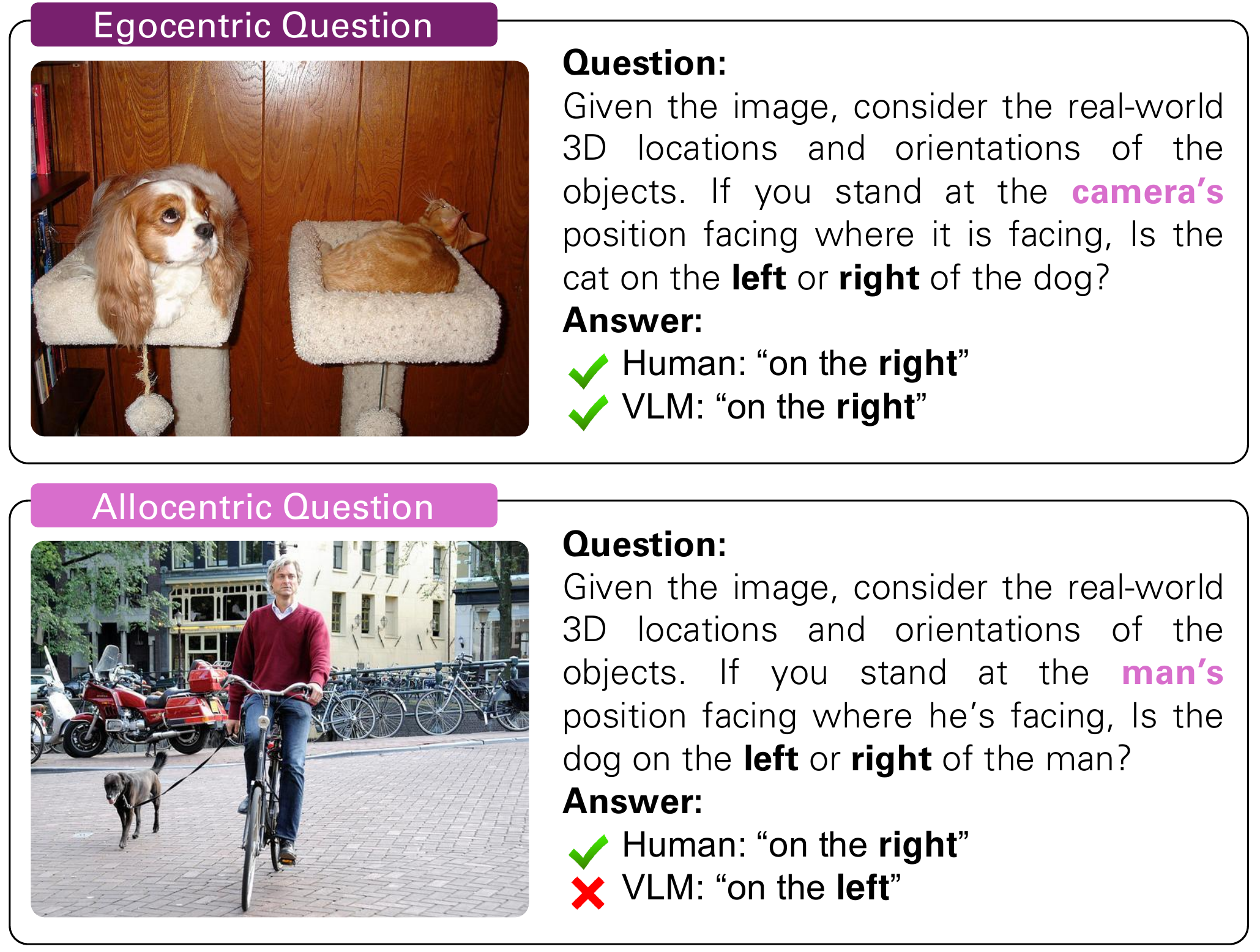}
  \vspace{-1.5\baselineskip}
  \caption{\textbf{Egocentric vs. Allocentric.} While VLMs perform well when questions are asked from an egocentric (\ie~camera's) perspective, they struggle when the same questions are posed from an allocentric perspective, showing a strong bias toward egocentric reasoning.}
  \label{fig:problem_definition}
  \vspace{-1.5\baselineskip}
\end{figure}

Despite these advances, progress remains largely confined to \emph{egocentric} spatial reasoning, and even the latest VLMs struggle with \emph{allocentric} reasoning---answering questions from perspectives other than the camera’s (Fig.~\ref{fig:problem_definition}). Allocentric reasoning is crucial for high-level planning, environmental interaction, and collaboration with autonomous agents~\cite{Fan:2022MineDojo,Yang:2023How2Com,Yang:2024Planning}. Moreover, it serves as a key benchmark for human-level spatial understanding. However, as analyzed by~\citet{zhang2024vision}, most VLMs exhibit a strong bias toward an \emph{egocentric} perspective. Even when explicitly prompted to adopt an \emph{allocentric} viewpoint, VLMs often revert to egocentric interpretations~\cite{linsley20243dpcbenchmarkvisualperspective, goral2024seeingeyesevaluatingvisual, zhang2024vision, zhang2024sphere}. Recent efforts to enhance spatial reasoning remain focused on improving egocentric reasoning~\cite{chen2024spatialvlm, cheng2024spatialrgpt, ma2024spatialpin}, leaving allocentric reasoning largely unaddressed.

To bridge the gap between VLMs and human perspective reasoning, we ask: \emph{What cognitive process allows humans to effortlessly shift perspectives?} Unlike current VLMs, humans seamlessly form internal representations of the physical world, making perspective reasoning an intuitive and natural process. The mechanism of creating internal representations, known as \emph{mental imagery}~\cite{Nanay2021,finke1989principles,Paivio:1979Imagery,Kosslyn:1978VisualImages,shepard1971mental}, plays a fundamental role in cognition, enabling us to simulate visual, spatial, and conceptual scenarios. This ability allows for abstraction beyond immediate perception, facilitating sophisticated spatial reasoning tasks such as mentally rotating objects, inferring occlusions, and envisioning alternative viewpoints~\cite{cole2022return, diwadkar1997viewpoint, burgess2006spatial}.

A key aspect of mental imagery is that it is not simply the process of visualizing a clear image from different perspectives; rather, it involves forming an \emph{abstract representation} of a scene that encodes essential spatial information and can be reinterpreted from a new perspective. From a computational standpoint, such an abstract representation is particularly advantageous, as equipping VLMs with the imaginative capability to generate novel views remains extremely challenging. In contrast, constructing an abstract representation requires significantly less computation and can be achieved procedurally.

Inspired by this, we introduce a novel framework for adapting perspectives in VLMs by simulating the mental imagery process and modifying the perspective in the given prompt. Our goal is to leverage the strengths of both VLMs and recent vision foundation models, such as object detection~\cite{liu2024grounding, Carion:2020DETR, Zhu:2021DDETR}, image segmentation~\cite{Kirillov2023:SAM, Ravi:2024SAM2}, and orientation estimation~\cite{wang2024orient}. The proposed framework takes an image and a perspective-based question as input and operates through three key stages. First, by simulating the mental imagery process, it builds an abstract representation of the scene in the input image. The VLM parses the prompt to identify objects in the image, while vision foundation models extract the center and orientation information of each object in 3D space. Second, the VLM analyzes the prompt to determine the reference object from whose perspective the question is asked, and transforms the abstraction to be aligned with that perspective. Finally, a new \emph{prompt} is generated by reinterpreting the scene from the reference object’s perspective. We explore different formats for rendering the scene abstraction: (1) a text-based representation, where objects are described using numerical 3D coordinates, and (2) an image-based representation, where objects are visualized as colored boxes corresponding to the original image. The newly generated prompt is fed to the VLM to obtain the final answer. Note that our framework is \emph{not} designed for a specific type of allocentric question. We leverage the egocentric reasoning capabilities of VLMs by performing allocentric-to-egocentric prompt conversion through scene abstraction, removing the perspective-related barrier in the question while preserving its original intent in the new prompt.

In our experiments on COMFORT++~\cite{zhang2024vision} and 3DSRBench~\cite{ma20243dsrbench}, our method achieves robust spatial reasoning across a variety of tasks and perspectives. In constrast, baseline VLMs and previous frameworks for spatial reasoning often struggle with even simple viewpoint shifts, reconfirming a notable bias toward the camera's perspective. These results highlight how our abstraction-based representation significantly enhances the spatial reasoning capabilities of VLMs beyond their default egocentric perspectives.
\section{Related Work}
\label{sec:related_work}

\vspace{-0.3\baselineskip}
\subsection{Spatial Reasoning with VLMs}
\label{subsec:related_spatial_reasoning}
\vspace{-0.3\baselineskip}
Building on the remarkable advancements of vision-language models (VLMs)~\cite{liu2023llava, liu2024llavanext, li2024llavaonevision, Qwen2.5-VL, deitke2024molmo}, recent studies have adapted VLMs for real-world spatial reasoning. Numerous evaluations revealed that VLMs struggle on even elementary spatial-perception tasks~\cite{ramakrishnan2024does, tang2024sparkle, wang2024picture, rahmanzadehgervi2024vision, fu2024blinkmultimodallargelanguage} and higher-level spatial reasoning based on images or videos~\cite{kamath2023s, liu2023visual, shiri2024empirical, song2024robospatial, yang2024thinking, li2024topviewrs}. SpatialVLM~\cite{chen2024spatialvlm} tackles this issue with a data-synthesis pipeline that injects rich spatial cues, while Cambrian-1~\cite{tong2024cambrian1fullyopenvisioncentric} introduces an architecture purposed for improved spatial reasoning. Another line of work allows VLMs to utilize richer vision-centric data such as points, depth maps or segmentation masks through fine-tuning~\cite{yuan2024robopoint, cai2024spatialbot, liu2025spatialcot, song2024robospatial} or employing auxiliary encoders~\cite{cheng2024spatialrgpt}. Taking a different approach, other works exploit the planning and programming abilities of language models, building LLM/VLM-in-the-loop systems that call external vision modules as needed~\cite{surismenon2023vipergpt, visprog, marsili2025visual}. Notably, SpatialPIN~\cite{ma2024spatialpin} extracts dense visual priors from multiple vision foundation models~\cite{jin2023perspective, liu2024one} and uses a VLM~\cite{hurst2024gpt} to combine and interpret this information.

\subsection{Visual Perspective-Taking}
\label{subsec:related_visual_perspective_taking}
Visual perspective-taking (VPT) is the ability to imagine an alternate viewpoint, whether from another person's perspective or a different camera angle. This ability is essential for fundamental human skills such as navigation, spatial awareness, and social interaction~\cite{psych_visual_persp, finke1989principles, Clark1991-CLAGIC, beckham2023visual}. To be regarded as a general vision agent capable of human-like reasoning, a VLM should possess robust perspective-taking abilities. However, recent analyses reveal that current VLMs fail to shift to allocentric perspectives, showing a strong bias toward the egocentric viewpoint of a given image~\cite{zhang2024vision, ma20243dsrbench, zhang2024sphere, linsley20243dpcbenchmarkvisualperspective, goral2024seeingeyesevaluatingvisual}. \citet{zhang2024vision} propose a synthetic evaluation protocol to assess whether VLMs can adopt different frames of reference (\ie~perspective). Likewise, 3DSRBench~\cite{ma20243dsrbench} includes real image-question pairs asked from an object's viewpoint, and finds that recent VLMs still demonstrate near chance level on perspective-related tasks. These findings suggest that while VLMs are rapidly improving in both complex visual reasoning~\cite{hurst2024gpt, team2024gemini, liu2023llava, Qwen2.5-VL, deitke2024molmo} and basic spatial reasoning~\cite{li2024llavaonevision, tong2024cambrian1fullyopenvisioncentric,chen2024spatialvlm, cai2024spatialbot, cheng2024spatialrgpt, heo2025omni}, their abilities remain confined to the egocentric viewpoint, posing a significant barrier to human-like reasoning. Recently, SAT~\cite{ray2024sat} proposed to improve VLMs' allocentric reasoning through instruction-tuning, yet it remains restricted to left/right relations with the need for annotations. In this work, we empower VLMs to reason from \emph{arbitrary} perspectives, by reformulating any spatial reasoning task into their default egocentric viewpoint, resulting in a generalizable framework.

\begin{figure}[b!]
  \centering
  \includegraphics[width=\linewidth]{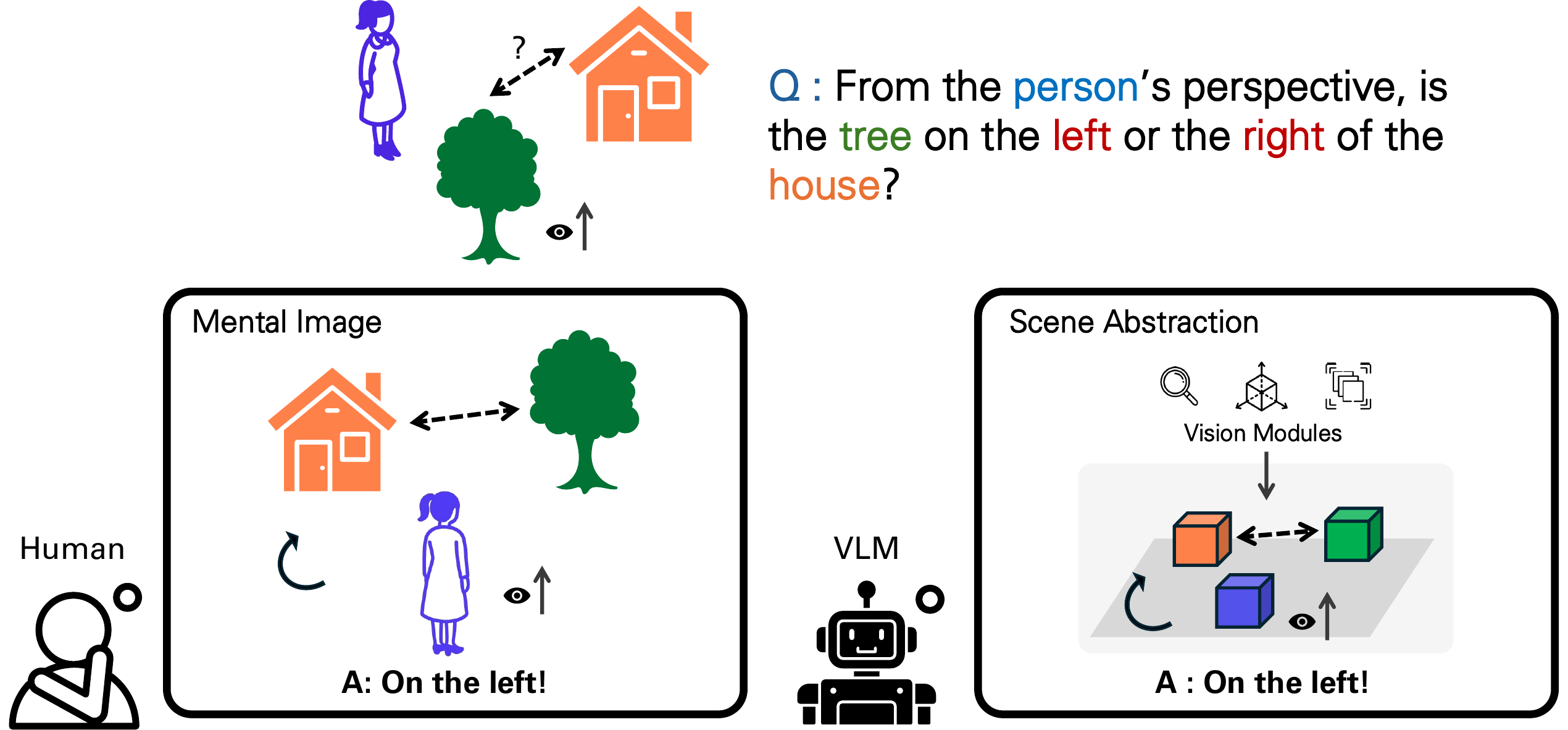}
  \vspace{-\baselineskip}
  \caption{\textbf{Mental Imagery Simulation.} Inspired by how humans employ mental imagery to reason from across different perspectives (left), we propose a similar process for VLMs, by constructing an explicit abstraction of the input scene and using it as a foundation for perspective changes (right).}
  \label{fig:concept}
\end{figure}

\subsection{Visual Prompting}
\label{subsec:related_visual_prompting}
Visual prompting frames an input image as an \emph{instruction} for a VLM, functioning similarly to how text prompts guide language models~\cite{wu2024visual, liu2024coarsecorrespondences, lei2024scaffolding, wu2024dettoolchain, yang2023fine, zhou2024image, shtedritski2023does}. Numerous studies have demonstrated its effectiveness by exploiting the inherent image comprehension capabilities of VLMs. Set-of-Marks~\cite{yang2023setofmarkpromptingunleashesextraordinary} augments each object in an image with its corresponding segmentation mask for more fine-grained visual grounding. Visual Sketchpad~\cite{hu2024visualsketchpadsketchingvisual} provides tool-based framework for VLMs to utilize drawing tools to annotate images for complex tasks such as math problem solving and visual search. 
Recent research further proposes \emph{visual} chain-of-thought (CoT) pipelines~\cite{li2025imaginereasoningspacemultimodal, rose2024visualchainthoughtbridging, zhou2024image, wu2024visualization, chen2024visual, shao2024visual, zhao2025cot} that visualize intermediate reasoning steps as images and feed them back to the model as auxiliary inputs. This visual feedback loop has proven effective for spatial tasks, as it anchors textual reasoning to concrete visual cues~\cite{li2025imaginereasoningspacemultimodal, wu2024visualization}. Building on this idea, we propose to transform an abstraction of a given scene and feed it back to a VLM in the form of a visual prompt, offering a new way for the model to reason from \emph{arbitrary} viewpoints.
\begin{figure*}[t!]
  \centering
  \includegraphics[width=\linewidth]{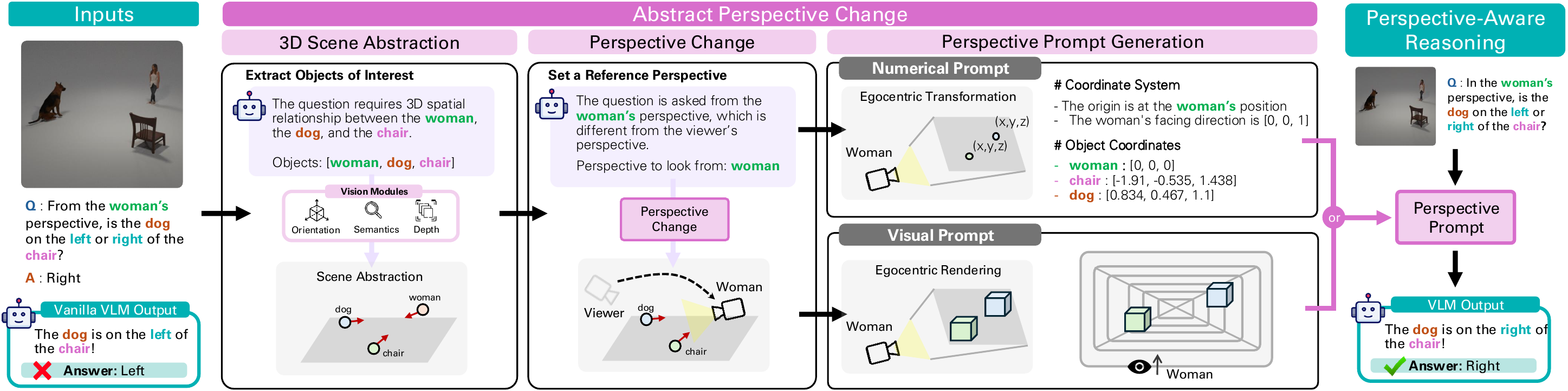}
  \vspace{-0.5\baselineskip}
  \caption{\textbf{Pipeline Overview of \methodname.} Our proposed framework consists of three stages. 1) Scene Abstraction (Sec.~\ref{subsec:scene_abstraction}): \methodname~first detects the objects of interest and build a coarse 3D abstraction of the scene using off-the-shelf vision foundation models. 2) Perspective Change (Sec.~\ref{subsec:perspective_change}): Then, a reference perspective is set and the abstraction is transformed into the reference viewer's egocentric coordinate frame. 3) Perspective Prompting (Sec.~\ref{subsec:perspective_prompting}): Finally, \methodname~passes the transformed scene to the VLM by producing (1) a numerical (textual) prompt or (2) an abstract visual prompt, and poses the question of interest from the reference perspective.}
  \label{fig:pipeline}
\end{figure*}

\vspace{-0.3\baselineskip}
\section{Method: Abstract Perspective Change}
\label{sec:method}
Our goal is to enable VLMs to solve spatial reasoning tasks from any given perspective (Fig.~\ref{fig:problem_definition}). Let us call the entity of the target perspective as the \emph{reference viewer}. Since VLMs inherently approach spatial reasoning from an egocentric perspective~\cite{zhang2024vision}, we propose to reformulate perspective-specific questions to align with the reference viewer's egocentric perspective. Inspired by theories in mental imagery~\cite{finke1989principles, Nanay2021, shepard1971mental}, we begin by explicitly building an \emph{abstraction} of the scene and use it as a foundation for shifting perspectives (Fig.~\ref{fig:concept}).

\vspace{-0.5\baselineskip}
\paragraph{Overview of \methodname.}
We call our approach Abstract Perspective Change  (\methodname), which consists of three main stages. (1) First, \methodname~constructs a coarse 3D abstraction of the scene from the input image by selecting and extracting objects of interest using off-the-shelf vision modules (Sec.~\ref{subsec:scene_abstraction}), drawing inspiration from human mental imagery~\cite{finke1989principles}. 
(2) Next, \methodname~selects a reference viewer for the spatial reasoning task among the objects of interests in the constructed scene abstraction. This determines ``where to look from". Such a formulation allows the conversion of the allocentric reasoning problem to an egocentric spatial reasoning task by performing a \emph{perspective change} that transforms the base coordinate system of the abstraction from the original camera view to that of the reference viewer (Sec.~\ref{subsec:perspective_change}). 
(3) Finally, the transformed abstracted scene, which can now be posed as an egocentric problem, is fed back into the VLM for spatial reasoning (Sec.~\ref{subsec:perspective_prompting}). We explore two alternative representations when providing the VLM with transformed astract scene information: 1) directly feeding numerical 3D coordinates of each object as a text prompt (numerical prompt), and 2) generating an abstract rendering of the scene as viewed by the reference perspective (visual prompt). An illustration of our \methodname~pipeline is shown in Fig.~\ref{fig:pipeline}, and we detail each step as follows.

\subsection{Scene Abstraction}
\label{subsec:scene_abstraction}
\methodname~begins by building a coarse 3D abstraction of the scene. Given an image $I$ and a spatial reasoning question $Q$, we define the abstraction of a scene as the set $S_E := \{ O_i \}^n_{i=1}$ composed of objects of interest from the question $Q$. Here, $E$ denotes that the abstraction is defined in the camera's egocentric coordinate system, and the number of objects of interest $n$ is determined by the VLM based on $Q$. Each $O_i$ corresponds to an object of interest in the image and is represented as a tuple $(t_i, c_i, p_i)$, where $t_i$ is the object's description, $c_i \in \mathbb{R}^3$ is its 3D position, and $p_i \in \mathbb{S}^3$ is a unit vector that indicates its orientation. Additionally, the camera is also included as an object of interest. This abstraction provides a minimal yet sufficient information in order to perform perspective changes, and mirrors how humans draw and rotate mental images when reasoning with perspectives~\cite{shepard1971mental, finke1989principles}. It allows for our \methodname~to convert an allocentric problem to an egocentric spatial reasoning task, which VLMs can better solve~\cite{zhang2024vision}. More details are described below.

\vspace{-0.5\baselineskip}
\paragraph{Extracting Objects of Interest.}
To determine which objects in the image should be included in the scene abstraction $S_E$, we provide the image $I$ and the question $Q$ to the VLM and instruct it to identify the list of objects necessary for answering the question. The VLM then returns the list of objects of interest, specified by their name, which we denote as $t_i$. The detailed instruction prompts are included in the \textbf{Appendix (Sec.~\ref{sec_app:prompt_details})}.

\vspace{-0.5\baselineskip}
\paragraph{Building Object Abstractions.}
Given the list of objects of interest, we complete our abstracted scene representation by extracting the position and orientation of each object $O_i$ using off-the-shelf vision foundation models. To obtain the 3D position of $O_i$, we first query GroundingDINO~\cite{liu2024grounding} with image $I$ and the object description $t_i$ and obtain its 2D bounding box $b_i$. We then crop $I$ with $b_i$, and utilize SAM~\cite{kirillov2023segment} to obtain a precise segmentation mask for $O_i$. Next, we extract the metric depth map of $I$ using DepthPro~\cite{bochkovskii2024depth} and unproject the pixels within the segmentation mask to 3D. Subsequently, the position $c_i$ is obtained by taking the median coordinate of this 3D point cloud. For further implementation details, please refer to the \textbf{Appendix (Sec.~\ref{sec_app:implementation_details})}.
Estimating the orientation $p_i$ for each object $O_i$ is also necessary to perform the desired perspective transformation. We utilize OrientAnything~\cite{wang2024orient}, which returns the object's frontal orientation within the camera coordinate system. For this, we crop the image with $b_i$ and feed the cropped image to OrientAnything to obtain $O_i$'s orientation, hence completing our scene abstraction representation.

\begin{figure*}[t!]
  \centering
  \includegraphics[width=\linewidth]{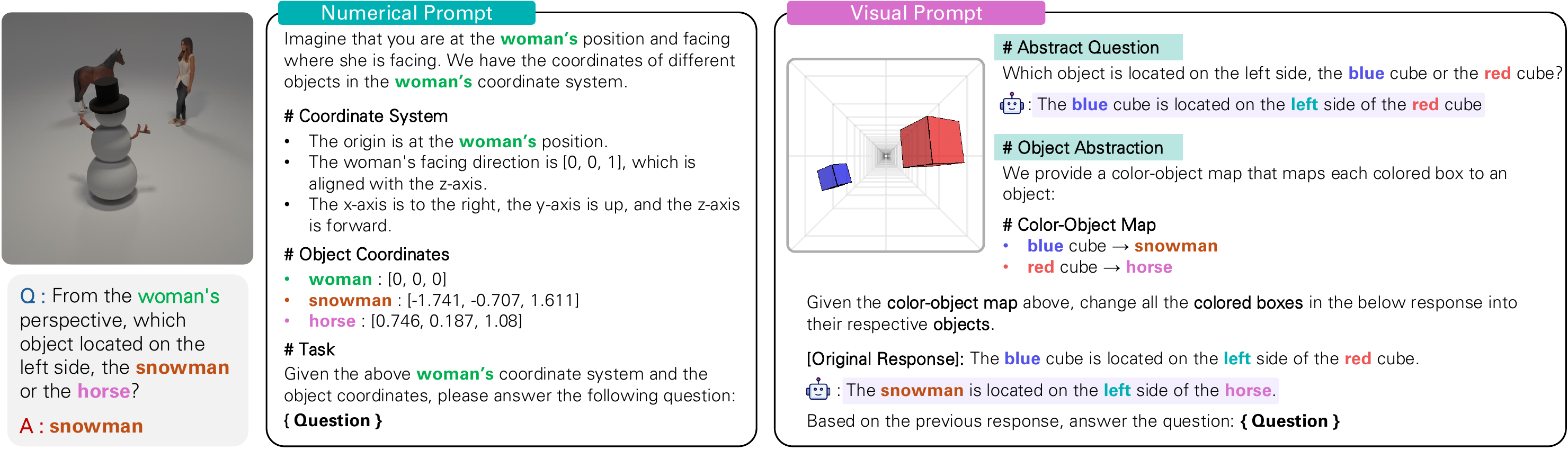}
  \vspace{-\baselineskip}
  \caption{\textbf{Perspective Prompt Samples.} We explore two variations of perspective prompting, numerical (left) and visual (right). Numerical (textual) prompt is generated by directly utilizing the 3D coordinate and orientation information. To generate the Visual prompt, we first place a colored cube at each object's identified 3D position then render the scene at the reference viewpoint, which results in an egocentric depiction of the scene. In addition, we construct an abstract question along with object-color mapping to ground the abstracted view.}
  \label{fig:perspective_prompting}
  \vspace{-0.75\baselineskip}
\end{figure*}

\subsection{Perspective Change}
\label{subsec:perspective_change}
With egocentric scene abstraction $S_E$ for a given image $I$ and question $Q$, \methodname~then determines the reference viewer and performs perspective change to obtain a transformed scene abstraction from the reference viewer’s perspective. This effectively converts an allocentric problem into an egocentric task, which VLMs find easier to handle.

\vspace{-0.5\baselineskip}
\paragraph{Setting a Reference Perspective.}
\methodname~ first determines ``where to look from" by selecting a reference viewer from the set of objects of interest. For this, we provide the spatial reasoning question $Q$ to the VLM and instruct it to identify the reference perspective from which the question should be answered. We denote the extracted reference perspective as $A$, and provide the complete instruction for perspective extraction in the \textbf{Appendix (Sec.~\ref{sec_app:prompt_details})}.

\vspace{-0.75\baselineskip}
\paragraph{Transforming Scene Abstraction.}
After identifying the reference viewer, we then transform the original camera-based scene abstraction $S_E$ into the reference viewer’s egocentric coordinate system. Specifically, we apply coordinate transformation from the camera’s frame to that of the reference viewer $A$. In the resulting abstraction $S_A$, the reference viewer $A$ is placed at the origin, and its orientation is aligned with the $z$-axis. This step supports \methodname’s main objective of reframing a general perspective question---typically an allocentric problem---into the reference viewer’s egocentric viewpoint, making it an egocentric task. Finally, we provide $S_A$ to the VLM so it can answer the question $Q$ from $A$’s perspective. We describe this stage more in depth below.

\subsection{Perspective Prompting}
\label{subsec:perspective_prompting}
The final step of \methodname~involves generating a prompt from the transformed scene abstraction $S_A$ to feed as input for the VLM. That is, how is the VLM asked with the transformed, now egocentric spatial reasoning task? We refer to our generated prompt as the \emph{perspective prompt} for image $I$ and question $Q$. Since VLMs can take images and text inputs, we explore two choices for the representation of this prompt: numerical (textual) and visual.

\vspace{-0.5\baselineskip}
\paragraph{Numerical (Textual) Prompt.}
Recall that an object abstraction in the transformed scene abstraction $S_A$ consists of the object's textual description, its corresponding 3D position, and its orientation, \ie~$O'_i = (t_i, c'_i, p'_i)$. Hence, a straightforward approach is to directly feed this information into the VLM. Specifically, we include the 3D position $c'_i$ in a predefined instruction template and instruct the VLM to directly solve the question $Q$. The full instruction template is provided in the \textbf{Appendix (Sec.~\ref{sec_app:prompt_details})}.

\vspace{-0.5\baselineskip}
\paragraph{Visual Prompt.}
Our goal is to let VLMs ``view the scene from $A$'s perspective"; thus an alternative choice for the perspective prompt is a visualization of our abstraction $S_A$. We begin by assigning each object an equal-sized cube, with each cube's position matching the objects' positions $c'_i$. We then render these cubes from the reference viewer $A$'s vantage point, generating an egocentric depiction of the scene abstraction. To distinguish between objects, each cube is assigned a unique color. When providing this information to the VLM, we modify the original question $Q$ to reflect the abstract visual representation. Specifically, we replace object names (\eg~\textit{``dog''}) with their corresponding colored cubes (\eg~\textit{``red box''}), forming an abstract question $Q^*$. Refer to Fig.~\ref{fig:perspective_prompting} for an example of an obtained abstraction question.
Putting it all together, the VLM receives as a prompt both the abstract rendered image---showing colored cubes---and the reformulated question $Q^*$. This allows it to answer spatial reasoning questions originally posed from \emph{arbitrary} perspectives by reasoning with this abstract, egocentric visual prompt. More details on the visual prompting process are presented in the \textbf{Appendix (Sec.~\ref{subsec_app:visual_prompt_rendering})}.
\begin{figure}[b!]
  \centering
  \includegraphics[width=\linewidth]{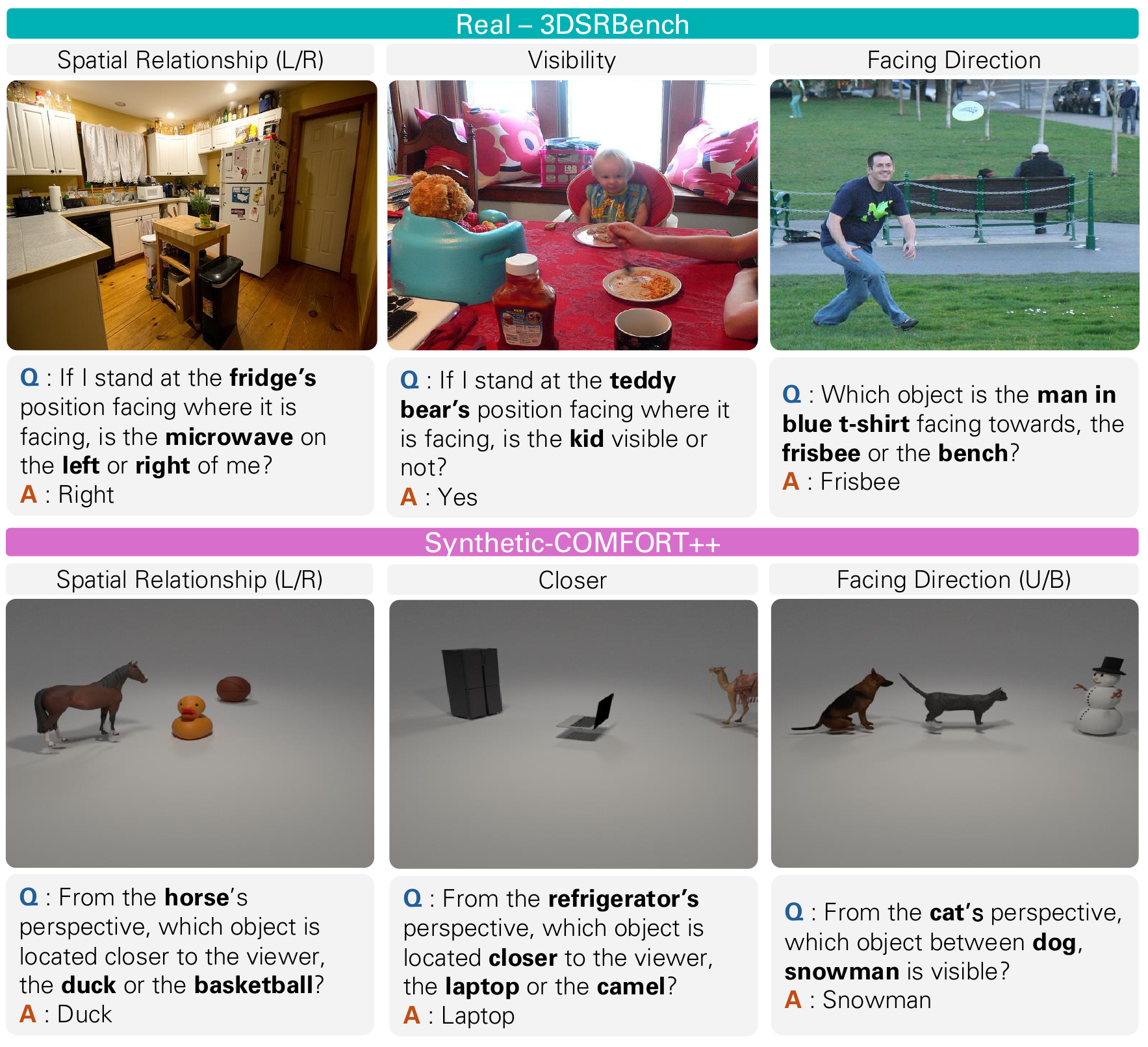}
  \vspace{-\baselineskip}
  \caption{\textbf{Benchmark Visualization.} Example image-question pairs from 3DSRBench~\cite{ma20243dsrbench} and COMFORT++~\cite{zhang2024vision} benchmarks. The tasks probe spatial reasoning across left-right relations, object visibility, closenss, and the facing direction of objects.}
  \label{fig:benchmark_examples}
\end{figure}

\section{Results}
\label{sec:results}
In this section, we present the experimental results of our \methodname~across a range of spatial reasoning tasks that include specified reference perspectives. We compare \methodname~to multiple baseline methods and show how our abstraction-based allocentric-to-egocentric reasoning framework enables the VLM to handle alternative perspectives. We use Qwen2.5-VL~\cite{Qwen2.5-VL} as our backbone VLM.

\begin{table*}[t!]
\centering
\begin{tabular*}{\textwidth}{@{\extracolsep{\fill}}lcccccccc@{}}
    \toprule
    \multirow{2}{*}{\textbf{Method}} & \multicolumn{4}{c}{\textbf{COMFORT++~\cite{zhang2024vision}}} & \multicolumn{3}{c}{\textbf{3DSRBench~\cite{ma20243dsrbench}}} \\
    \cmidrule(lr){2-5} \cmidrule(lr){6-8}
    & \textbf{left/right} & \textbf{closer} & \textbf{visibility} & \textbf{facing} 
    & \textbf{left/right} & \textbf{visibility} & \textbf{facing} \\
    \midrule
    Random & 50.00 & 50.00 & 50.00 & 50.00 & 50.00 & 50.00 & 50.00 \\
    \midrule
    \cellcolor{blue!10}LLaVA-NeXT~\cite{liu2024llavanext} & 48.00 & 47.33 & 40.00 & 39.00 & 34.10 & 41.57 & 50.29 \\ 
    \cellcolor{blue!10}LLaVA-OneVision~\cite{li2024llavaonevision} & 55.33 & 79.00 & 50.94 & 38.33 & 32.09 & 46.51 & 60.12 \\ 
    \cellcolor{blue!10}Molmo~\cite{deitke2024molmo} & 36.33 & 35.67 & 31.88 & 29.00 & 19.77 & 22.97 & 32.08 \\ 
    \cellcolor{blue!10}Qwen2.5-VL~\cite{Qwen2.5-VL} & 43.33 & 74.33 & 51.25 & 43.00 & 34.96 & 45.06 & 53.47 \\ 
    \cellcolor{blue!10}Cambrian-1~\cite{tong2024cambrian1fullyopenvisioncentric} & 52.00 & 79.00 & 57.50 & 41.00 & 40.97 & 49.71 & \underline{65.03} \\ 
    \cellcolor{blue!10}GPT-4o~\cite{hurst2024gpt} & 41.00 & 61.33 & 53.75 & 38.67 & 2.01 & 40.12 & 47.70 \\ 
    \cellcolor{blue!10}Gemini-2.0-Flash~\cite{team2024gemini} & 43.67 & 26.00 & 40.31 & 13.00 & 24.93 & 57.65 & 55.20 \\ 
    \midrule
    \cellcolor{ForestGreen!15}SpatialVLM~\cite{chen2024spatialvlm} & 46.00 & 41.67 & 42.81 & 29.33 & 22.35 & 46.51 & 47.11 \\ 
    \cellcolor{ForestGreen!15}SpatialRGPT~\cite{cheng2024spatialrgpt} & 27.08 & 33.90 & 29.25 & 1.33 & 25.98 & 27.19 & 42.55 \\ 
    \cellcolor{ForestGreen!15}SpatialPIN~\cite{ma2024spatialpin} & 19.62 & 23.96 & 48.43 & 43.91 & 11.10 & 42.40 & 11.66 \\
    \midrule
    \cellcolor{RubineRed!15}SpatialPIN$^*$~\cite{ma2024spatialpin} & 59.80 & 70.45 & 49.84 & 50.51 & 50.10 & 52.30 & 28.86 \\
    \cellcolor{RubineRed!15}ViewCrafter~\cite{yu2024viewcrafter} & 32.33 & 53.00 & 38.75 & 37.46 & 28.41 & 22.47 & 18.31 \\ 
    \midrule
    \cellcolor{Gray!15}\textbf{\methodname\texttt{-Num}~(Ours)} & \underline{88.67} & \textbf{96.00} & \underline{71.25} & \underline{62.00} & \underline{71.92} & \underline{62.79} & 60.98 \\
    \cellcolor{Gray!15}\textbf{\methodname\texttt{-Vis}~(Ours)} & \textbf{89.67} & \underline{94.33} & \textbf{90.00} & \textbf{88.33} & \textbf{72.78} & \textbf{67.44} & \textbf{66.47} \\
    \bottomrule
\end{tabular*}
\caption{\textbf{Quantitative Comparisons.} Purple (\colorbox{blue!10}{\phantom{a}}) represents \emph{pure VLMs}, green (\colorbox{ForestGreen!15}{\phantom{a}}) represents \emph{grounded VLMs}, and red (\colorbox{RubineRed!15}{\phantom{a}}) represents \emph{dense reconstruction-based} frameworks. Gray (\colorbox{Gray!15}{\phantom{a}}) corresponds to our \methodname. \textbf{Bold} and \underline{underline} indicate the best and the second-best result for each column, respectively. \methodname\texttt{-Num} and \methodname\texttt{-Vis} refer to our method employing numerical prompt and visual prompt, respectively.}
\label{tab:main_quantitative}
\end{table*}


\subsection{Evaluation Settings}
\label{subsec:evaluation_settings}

\paragraph{Benchmarks.}
We validate our \methodname~on both synthetic~\cite{zhang2024vision} and real-world~\cite{ma20243dsrbench} benchmarks in which the spatial reasoning requires perspective changes. Sample image-question pairs from each benchmark are shown in Fig.~\ref{fig:benchmark_examples}.

\begin{itemize}
    \item \textbf{COMFORT++:} Zhang \etal~\cite{zhang2024vision} introduce COMFORT, a benchmark synthesis protocol designed to evaluate VLMs on perspective-aware spatial reasoning. It employs a simple Blender~\cite{blender} rendering pipeline to place multiple objects in a synthetic scene with one reference viewer and various other objects. Each scene poses a spatial reasoning question from the reference viewer's perspective. Building on COMFORT, we construct four types of spatial reasoning tasks that require a reference viewer different from the camera: \textit{left/right}, \textit{closer/further}, \textit{visibility}, and \textit{facing}.
    \item \textbf{3DSRBench:} Ma \etal~\cite{ma20243dsrbench} introduce a 3D spatial reasoning benchmark based on MS-COCO images~\cite{lin2014microsoft}. We focus on three categories that require an allocentric viewpoint: \textit{left/right}, \textit{visibility}, and \textit{facing}. Note that we recast the original \emph{front/behind} question in 3DSRBench into a \emph{visibility} question using the same images. We provide further discussion on the dataset and the evaluation protocol in the \textbf{Appendix (Sec.~\ref{sec_app:evaluation_details})}.
\end{itemize}

\vspace{-\baselineskip}
\paragraph{Baselines: VLMs.}
We benchmark our \methodname~against multiple state-of-the-art VLMs, including both open-source and proprietary models. For open-source, we include LLaVA-NeXT~\cite{liu2024llavanext}, LLaVA-OneVision~\cite{li2024llavaonevision}, Molmo~\cite{deitke2024molmo}, and Qwen2.5-VL~\cite{Qwen2.5-VL}. We also include proprietary models: GPT-4o~\cite{hurst2024gpt} and Gemini-2.0-Flash~\cite{team2024gemini}. We refer to these as \emph{pure VLMs}. 
Additionally, we compare against \emph{grounded VLMs}, which include models explicitly tuned for spatial reasoning, such as SpatialVLM~\cite{chen2024spatialvlm} and SpatialRGPT~\cite{cheng2024spatialrgpt}. We also include SpatialPIN~\cite{ma2024spatialpin}, which leverages interactions between VLMs and vision foundation models for complex spatial reasoning.

\vspace{-0.3\baselineskip}
\paragraph{Baselines: Dense Reconstruction.}
To compare \methodname~with standard dense reconstruction techniques for novel view synthesis, we introduce two baselines. First, we extend SpatialPIN~\cite{ma2024spatialpin} to include our perspective change phase (Sec.~\ref{subsec:perspective_change}). We use the generated meshes from its original pipeline and render the meshes from the reference perspective, and denote this extension as SpatialPIN$^*$. Refer to the \textbf{Appendix (Sec.~\ref{sec_app:analysis_dense_recon}}) for more details. Second, we adopt ViewCrafter~\cite{yu2024viewcrafter}, a novel view synthesis method designed for single-image inputs. For both baselines, we synthesize a novel view according to the reference perspective's relative pose, and feed the resulting image to the VLM for spatial reasoning.

\subsection{Evaluation on COMFORT++~\cite{zhang2024vision}}
\label{subsec:results_comfort}
Tab.~\ref{tab:main_quantitative} (cols 2-5) provides quantitative comparisons on COMFORT++. Here, \methodname\texttt{-Vis} refers to our visual prompt, and \methodname\texttt{-Num} corresponds to the numerical prompt. 
Even though the benchmark consists of objects rendered in a simple, synthetic scene (see Fig.~\ref{fig:benchmark_examples}), we find that most pure VLMs (rows 3-9) struggle with the \emph{left/right} task, hovering around chance level with the best performing model LLaVA-OneVision scoring only 55.33\%. This confirms earlier observations~\cite{zhang2024vision} that VLMs fail to adopt alternative perspectives. Even specialist VLMs designed for spatial reasoning perform poorly, with SpatialVLM at 46.0\% and both SpatialRGPT and SpatialPIN also exhibiting low accuracy.
We observed that SpatialRGPT often generates hallucinated responses unrelated to the instruction, thereby resulting in low accuracy. While SpatialPIN$^*$---employing perspective change---shows better performance, low-quality meshes often bottleneck further improvements (refer to Sec.~\ref{sec_app:analysis_dense_recon} for more discussions). In contrast, our \methodname~significantly outperforms these baselines, achieving 89.67\% accuracy with a visual prompt and 88.67\% with a numerical (textual) prompt.

For the \emph{closer} task, some VLMs show relatively high accuracy (79.00\% for both LLaVA-OneVision and Cambrian-1), likely since they can also solve the question by comparing object distances directly from the egocentric viewpoint. Even in this case, \methodname~achieves higher accuracy, attaining 96\% when using a numerical prompt. Moreover, for \emph{visibility} and \emph{facing} categories, the baseline models perform at near-chance levels, failing to take the reference perspectives into account. Notably, \methodname~exhibits a performance gap between visual and numerical prompts, with the visual prompt outperforming the numerical one by +18.75\% and +26.33\%, respectively. We attribute this difference to trivial logical errors that VLMs often make when relying on numerical coordinates. In contrast, for these two tasks the visual prompt requires only simple visual perception, mitigating such logical errors and achieving more accurate results.

\begin{figure*}[t!]
  \centering
  \includegraphics[width=\linewidth]{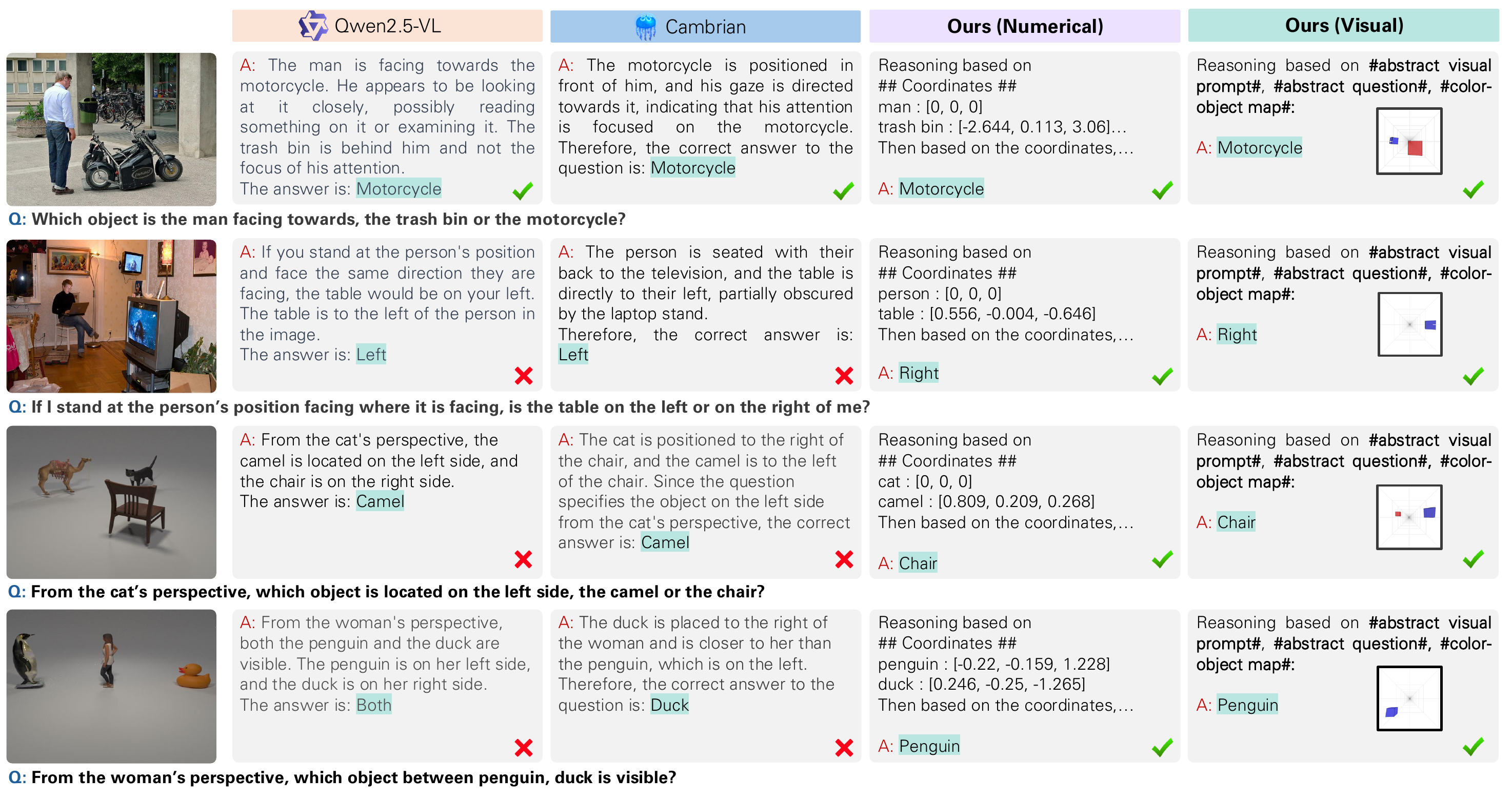}
  \vspace{-1.5\baselineskip}
  \caption{\textbf{Spatial Reasoning with Perspective Change.} Recent VLMs such as Qwen2.5-VL~\cite{Qwen2.5-VL} and Cambrian-1~\cite{tong2024cambrian1fullyopenvisioncentric} often struggle with spatial reasoning tasks that require a shift to a specific reference viewpoint. In constrast, our \methodname~effectively handles such perspective changes by constructing a scene abstraction and delivering the transformed view through a simple prompting technique.}
  \label{fig:qualiative}
  \vspace{-1.25\baselineskip}
\end{figure*}

\subsection{Evaluation on 3DSRBench~\cite{ma20243dsrbench}}
Tab.~\ref{tab:main_quantitative} (cols 6-8) presents quantitative comparisons on 3DSRBench, which includes real images. Compared to the synthetic environment in COMFORT++, using real images introduces additional noise into both the VLMs' visual reasoning and our \methodname's scene abstraction phase. For the \emph{left/right} task, baseline VLMs consistently fall under 50\% accuracy, including the grounded VLMs. Even SpatialPIN$^*$ with perspective change only reaches 50.10\%, at near chance-level. We find that using ViewCrafter to generate a novel view from the reference perspective yields 28.41\% accuracy, due to the noise and hallucinations during generation. We provide visualizations of the rendered views from both SpatialPIN$^*$~\cite{ma2024spatialpin} and ViewCrafter~\cite{yu2024viewcrafter} in the \textbf{Appendix (Sec.~\ref{sec_app:analysis_dense_recon})}, along with more discussions on the different between our abstraction-based approach and the dense reconstruction-based approaches. Compared to other baselines, \methodname~consistently achieves accuracies above 60\% both with visual and numerical prompts, showing that our framework is robust to real images.

For \emph{visibility} task, while our method outperforms the baselines, the accuracy appears lower than previous tasks---67.44\% with the visual prompt and 62.79\% with the numerical prompt. We attribute this decline to the noise in the scene abstraction phase, particularly errors in detected orientations or centroids. This issue is evident when compared to the \emph{visibility} task in COMFORT++, which employs the same question format but with a simpler scene. Lastly, for the \emph{facing} task, Cambrian-1 achieves 64.03\%, yet our \methodname~with a visual prompt still leads at 66.47\%. Qualitative examples on 3DSRBench are shown in Fig.~\ref{fig:qualiative}.

\begin{figure}[b!]
  \centering
  \includegraphics[width=\linewidth]{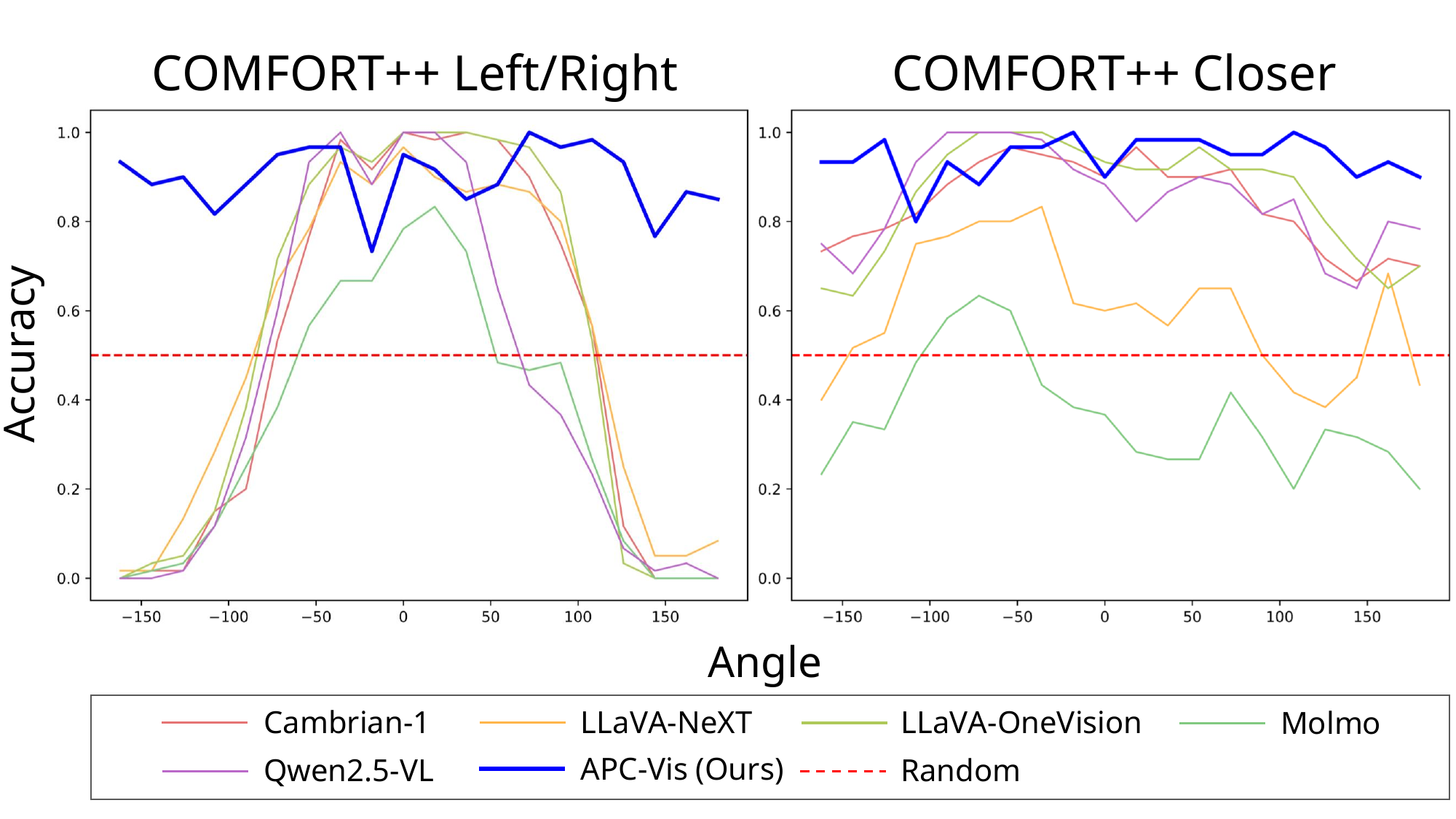}
  \vspace{-1.5\baselineskip}
  \caption{\textbf{Perspective Awareness.} Each plot shows accuracy versus the angular offset $\theta$ between the camera and the reference viewpoint. While baselines show clear degradation at certain ranges of $\theta$, \methodname~retains robust accuracy across all angles, demonstrating strong perspective-aware reasoning.}
  \vspace{-\baselineskip}
  \label{fig:angle_wise}
\end{figure}

\vspace{-0.3\baselineskip}
\subsection{Probing the Perspective Awareness of VLMs}
\label{subsec:results_persp_awareness}
\vspace{-0.3\baselineskip}
Finally, we analyze the \emph{perspective-awareness} of each method by assessing spatial reasoning accuracy across different viewpoints. Specifically, we select two tasks---\emph{left/right} and \emph{closer}---and construct 60 scenes similar following our setting in COMFORT++. Each scene is rendered from 20 evenly spaced azimuth angles. We define $\theta$ as the angular offset between the camera's orientation and the reference viewer's orientation. Here, for $\theta=0^\circ$ the camera is aligned with the reference perspective, while $\theta=180^\circ$ indicates that the reference viewer is facing towards the camera.

The results are shown in Fig.~\ref{fig:angle_wise}. For the \emph{left/right} task (left), the baselines exhibit clear bell-shaped curves, achieving near perfect accuracy when $\theta$ is close to $0^\circ$ (egocentric) but rapidly declining as the magnitude of $\theta$ increases (allocentric). In contrast, \methodname~maintains consistently high accuracy across all angles, demonstrating strong \emph{perspective-aware} reasoning. For the \emph{closer} task (right), baseline models also show noticeable accuracy drops, especially near the leftmost and rightmost $\theta$ ranges. \methodname~consistently achieves over 80\% accuracy, robustly handling viewpoints regardless of their deviation from the egocentric perspective.
\vspace{-0.5\baselineskip}
\section{Conclusion}
\label{sec:conclusion}
\vspace{-0.3\baselineskip}
In this work, we introduced \methodname, a framework empowering VLMs with the capability of perspective-aware reasoning. Our key idea is to simulate the mental imagery process of humans, abstracting the scene in an image to facilitate allocentric-to-egocentric perspective shifts, and in turn convey the transformed view to the VLM in the form of a prompt. The scene abstraction is constructed using vision foundation models for object detection, segmentation, and orientation estimation. The reframed prompt from the new perspective, either in text or image form, is then processed by VLMs, leveraging their egocentric reasoning capabilities. As shown by our experiment on both synthetic and real spatial reasoning benchmarks, \methodname~enables robust accuracy across diverse perspectives, thereby opening new possibilities of VLMs on real-world spatial tasks.

{
    \small
    \bibliographystyle{ieeenat_fullname}
    \bibliography{main}
}

\newpage
\appendix

\renewcommand{\thesection}{\Alph{section}}
\clearpage
\newpage

\section*{Appendix}

\noindent
In this appendix, we first discuss the limitations of our work and potential directions for future work (Sec.~\ref{sec_app:limitations_future_work}). We then analyze the failure cases of two dense reconstruction-based baselines---SpatialPIN$^*$~\cite{ma2024spatialpin} and ViewCrafter~\cite{yu2024viewcrafter}---in allocentric reasoning scenarios (Sec.~\ref{sec_app:analysis_dense_recon}). We describe the implementation details of our \methodname~framework (Sec.~\ref{sec_app:implementation_details}) and provide details on the evaluation setups (Sec.~\ref{sec_app:evaluation_details}). Finally, we provide the text prompts used in each stage of our method (Sec.~\ref{sec_app:prompt_details}).

\section{Limitations and Future Work}
\label{sec_app:limitations_future_work}
Our \methodname~framework empowers VLMs with perspective-aware spatial reasoning, but its use of multiple vision foundation models~\cite{liu2024grounding, kirillov2023segment, Wang:2024OrientAnything} introduces additional memory usage compared with running the VLM alone. In our experiments, we ran inference on two NVIDIA RTX 3090 GPUs each with 24GB VRAM.

While in this work we introduced a minimal yet effective form of 3D abstraction for perspective change in VLMs, exploring richer scene abstractions from images could offer an promising direction for future research---such as the use of 3D bounding boxes~\cite{nie2020total3dunderstanding, brazil2023omni3d, yao2024open} and coarse, semantic 3D scene reconstructions~\cite{dahnert2021panoptic, cao2022monoscene, nie2023learning}.
\section{Analysis on Dense Reconstruction Baselines}
\label{sec_app:analysis_dense_recon}

\begin{figure*}[b]
  \centering
  \includegraphics[width=\linewidth]{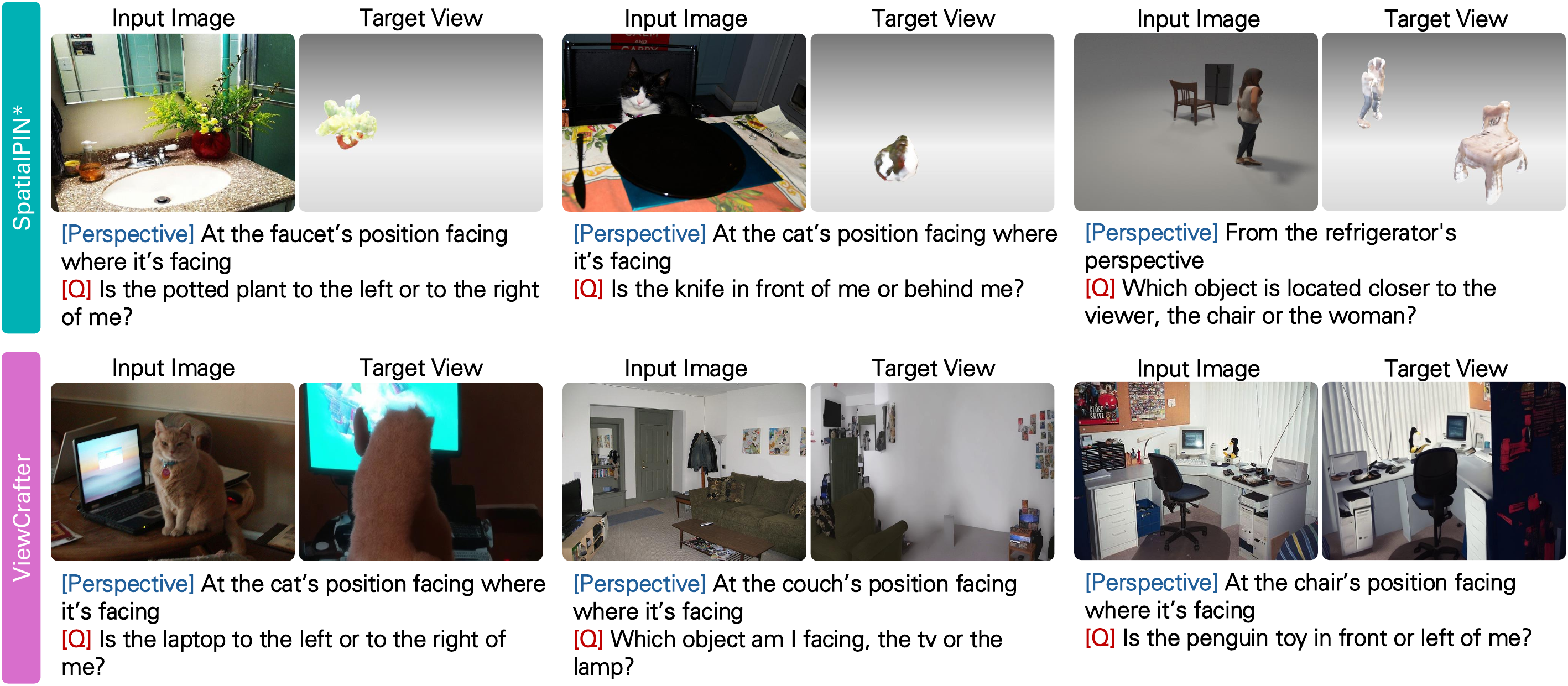}
  \caption{\textbf{Dense Reconstruction Baseline Examples.} Novel views synthesized by SpatialPIN$^*$~\cite{ma2024spatialpin} and ViewCrafter~\cite{yu2024viewcrafter} both display noisy and inaccurate objects and scene structures lacking the original context of the input image, thereby leading to low accuracy when VLMs are fed the images as a visual input for spatial reasoning.}
  \label{fig:dense_recon_examples}
\end{figure*}

In this section, we further discuss the dense reconstruction-based baselines introduced in Sec.~\ref{subsec:evaluation_settings}. In contrast to \methodname's abstraction-based approach, another intuitive approach for perspective-aware spatial reasoning is to perform a dense 3D reconstruction of the scene and then render a novel view from the target perspective. This new view can then be provided to the VLM instead of the visual prompt used in Sec.~\ref{subsec:perspective_prompting}. We explore two such approaches that involve dense 3D reconstruction process: (1) a modified version of SpatialPIN~\cite{ma2024spatialpin}, which directly lifts objects from the image into meshes and renders them from the target view, and (2) ViewCrafter~\cite{yu2024viewcrafter}, which synthesizes novel views by using an intermediate point cloud reconstruction. As the original SpatialPIN~\cite{ma2024spatialpin} does not include a rendering phase for novel target perspectives, we refer to our extended pipeline as SpatialPIN$^*$. For the inference of SpatialPIN$^*$, we used One-2-3-45~\cite{liu2023one} in contrast to One-2-3-45++~\cite{liu2024one} in the original paper due to the limited access of the API.

\begin{table}[h!]
\centering
\footnotesize
\renewcommand{\arraystretch}{1.0}
\begin{tabularx}{0.95\linewidth}{l c c c}
    \toprule
    Method & SpatialPIN$^*$~\cite{ma2024spatialpin} & ViewCrafter~\cite{yu2024viewcrafter} & \methodname~(Ours) \\
    \midrule
    Time (s) & 336.21 & 260.57 & 17.47 \\
    \bottomrule
\end{tabularx}
\caption{\textbf{Inference Time Comparison.} Both dense reconstruction-based baselines~\cite{ma2024spatialpin, yu2024viewcrafter} require over 14 times the inference time of our \methodname~to answer a single question.}
\label{tab:inference_time}
\end{table}

\noindent
While a dense reconstruction-based approach may appear to be an obvious alternative to our abstraction-based framework, our experiments show that constructing an accurate and descriptive view of the target perspective is challenging and expensive. As illustrated in Fig.~\ref{fig:dense_recon_examples}, the synthesized novel views from both SpatialPIN$^*$ (row 1) and ViewCrafter (row 2) are often excessively noisy and fail to preserve the context of the input image. Consequently, providing these reconstructed views to the VLM for spatial reasoning results in lower accuracy as previously shown in Tab.~\ref{tab:main_quantitative}. In addition, both methods incur notably longer inference times due to the dense 3D reconstruction steps, as shown in Tab.~\ref{tab:inference_time}. In contrast, as in our \methodname, constructing an minimal abstraction of the scene with precise mappings between the original objects and their abstractions not only yields more accurate reasoning but also substantially reduces inference time.
\renewcommand{\thefootnote}{\arabic{footnote}}

\section{Implementation Details}
\label{sec_app:implementation_details}
In this section, we provide the implementation details of our \methodname~framework in Sec.~\ref{sec:method}. As the backbone VLM, we used \texttt{Qwen2.5-VL-7B-Instruct}\footnote{\url{https://huggingface.co/Qwen/Qwen2.5-VL-7B-Instruct}}.

\subsection{Scene Abstraction}
\label{subsec_app:scene_abstraction}

\begin{figure}[h!]
  \centering
  \includegraphics[width=\linewidth]{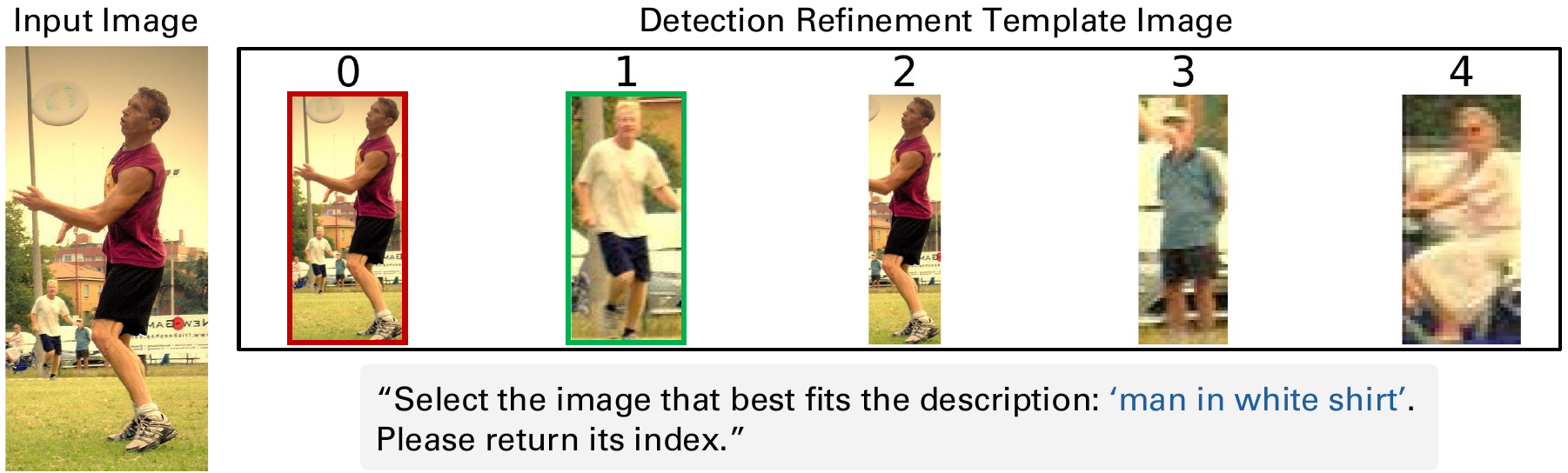}
  \vspace{-\baselineskip}
  \caption{\textbf{Detection Refinement Example.} Starting with candidate detections from GroundingDINO~\cite{liu2024grounding}, we select the top $k$ predictions and present them as a grid of cropped images (right). We then query the VLM to return the index that best aligns with the input text prompt. \textcolor{red}{Red} indicates GroundingDINO's initial choice and \textcolor{teal}{green} indicates the refined choice.}
  \label{fig:detection_refine}
  \vspace{-1.5\baselineskip}
\end{figure}

\paragraph{Detection Refinement with VLM.}
While GroundingDINO~\cite{liu2024grounding} excels in object detection, it often struggles when the input text prompt is complex. We add a simple refinement stage utilizing the VLM for improved detection accuracy. For each object description $t_i$ we keep GroundingDINO's predicted candidates whose confidence exceeds a threshold $s$, then select the top $k$ candidates. The corresponding image crops are laid out in a grid, and we query the VLM to select the crop that best matched $t_i$. We set $s=0.15$ and $k=5$. Fig.~\ref{fig:detection_refine} illustrates a case in which the initial GroundingDINO output is incorrect but is corrected by this refinement step.

\paragraph{Filtering Outliers.}
To obtain the 3D position of each object abstraction $O_i \in S_E$, we unproject the segmented pixels using the predicted depth map. To handle outliers caused by background pixels being included in the segmentation masks, we filter out the points whose depth values fall outside the range $[ 0.9d_i, 1.1d_i ]$, where $d_i$ is the mode depth within the mask. We then assign the coordinate-wise median of the remaining points in the remaining points as the 3D position $c_i$ of object $O_i$.

\subsection{Egocentric Rephrasing}
\label{subsec_app:egocentric_rephrasing}
Recall that our \methodname~converts an \emph{allocentric} question $Q$---originally stated with respect to a reference viewpoint $A$---into an \emph{egocentric} one posed from $A$ itself. To ensure compatibility with the perspective prompts introduced in Sec.~\ref{subsec:perspective_prompting}, we remove the explicit perspective descriptions from $Q$. In practice, we query the VLM to rewrite $Q$, excluding the phrases that mention a reference perspective. In turn we obtain a perspective-agnostic reformulation of the task, which is then used in each type of perspective prompt.

\subsection{Visual Prompt Rendering.}
\label{subsec_app:visual_prompt_rendering}
To render a \emph{visual prompt} from the transformed scene abstraction $S_A = \{ O'_i \}^{n}_{i=1}$ as shown in Fig.~\ref{fig:perspective_prompting}, we use the Trimesh renderer~\cite{trimesh}. Note that $S_A$ is defined in the coordinate system of the reference perspective $A$. Each object $O'_i$ is converted to an equal-sized cube with distinct colors, and the visual prompt is obtained by rendering the scene accordingly. Given the camera in $S_A$ faces in the positive $z$-direction, only the objects with $z > 0$ appear in the visual prompt. Objects with $z \leq 0$ are considered to be out of view (\ie~not visible) from perspective $A$.

\paragraph{Normalization.}
To prevent cubes from appearing too small or large in the visual prompt, we normalize the coordinates of $S_A$, ensuring $z$ values lie within a predefined range $[z_{\text{min}}, z_{\text{max}}]$. Likewise, we scale the $x, y$ coordinates into a fixed range $[-d^*, d^*]$ to keep objects within the view frustum.

\paragraph{Camera Translation.}
By default, we place the camera at reference viewer's position---the origin of $S_A$. As an exception, for the \emph{left/right} task in 3DSRBench~\cite{ma20243dsrbench}, we shift the camera backward along the $z$-axis to ensure all objects in the scene appear in the visual prompt. This adjustment is applied to match the benchmark's setup, where an object that lies \emph{behind} and \emph{to the right} of a reference viewer is still treated as being on the right side from that viewer's perspective.

\section{Evaluation Details}
\label{sec_app:evaluation_details}
In this section, we provide further details on our evaluation setup in Sec.~\ref{sec:results}. Each VLM response is scored with a two-step process that combines \emph{exact matching} and \emph{LLM-assisted} evaluation. We first perform exact matching: if the response consists solely of the correct option index or the exact answer phrase, we label it as correct. Otherwise, we pass the entire response to an LLM along with the answer to determine its correctness. For this, we used the judgment prompt template from VLMEvalKit~\cite{duan2024vlmevalkit}.

Following 3DSRBench~\cite{ma20243dsrbench}, we employ CircularEval~\cite{liu2024mmbenchmultimodalmodelallaround}, which takes into account VLM's response consistency by permuting the answer options for each image-question pair. The VLM is considered to be correct for a question $Q$ only if it selects the correct across all permutations. CircularEval is applied for both COMFORT++~\cite{zhang2024vision} and 3DSRBench~\cite{ma20243dsrbench}.

To construct the COMFORT++ benchmark for each task, we first collected 7 object meshes from the original implementation~\cite{zhang2024vision} and additional 6 meshes from Objaverse-XL~\cite{deitke2023objaverse}. 
For the \emph{left/right} and \emph{closer} tasks, we arranged three objects in a predefined layout, designating one as the reference viewer, and added random perturbations to the objects' $x, y$ coordinates to diversify the layouts. We prepared 60 scenes and rendered each from 20 evenly spaced azimuth viewpoints. Then, we randomly sampled five views per scene, resulting in a total of 300 images for each task. 
For the \emph{visibility} task, we created 160 scenes, each containing a reference viewer and single target object positioned so that the object is either visible or invisible from the viewer's perspective. We rendered each scene two opposite viewpoints, yielding 320 images. Finally, for the \emph{facing} task, we arranged three objects in a linear configuration, setting the central object as the reference viewer, and oriented it to face either one of the two remaining objects. Each scene is rendered once, resulting in 300 images in total.

For 3DSRBench~\cite{ma20243dsrbench}, we used the original \emph{left/right} and \emph{facing} criteria. We recasted the \emph{front/behind} task as a \emph{visibility} judgment for two reasons: (i) the provided task can be more naturally interpreted as deciding whether an object is visible from the reference object’s viewpoint, and (ii) VLMs struggle to infer that an object is \emph{behind} it when the object is not present in the image itself. This adjustment better serves our goal of measuring the egocentric and allocentric reasoning capabilities of VLMs.
\section{Details on Text Prompts}
\label{sec_app:prompt_details}
In this section, we present the text prompts used at each stage of our \methodname~pipeline. To guide the VLM towards the desired response format, we include examplar question-answer pairs for in-context learning. For the text prompt fed along with the visual prompt, we add simple prompt engineering to help suppress hallucinations: we (i) define the the relation \textit{``facing towards''} and (ii) explicitly that the larger object is considered as being closer to the viewer---an assumption that holds since our abstraction assigns equal size to every object.

\vspace{\baselineskip}
\noindent
\textbf{(1) Scene Abstraction (Sec.~\ref{subsec:scene_abstraction})} --- Extracting Objects of Interest.

\begin{tcolorbox}[
  breakable,
  width=\linewidth,
  colback=blue!3!white,
  colframe=blue!60!black,
  boxrule=0.6pt,
  arc=1mm,
  left=1.2mm,
  right=1.2mm,
  top=0.8mm,
  bottom=0.8mm,
]

\ttfamily\small

\textbf{\# Situation Description}\par
Given an image and a spatial‑reasoning question, identify \emph{all} entities mentioned in the question.\par\medskip

\textbf{\# Example}\par
[Question] You are standing at the airplane’s position, facing where it is facing.  
Is the person on your left or right?\par
[Detect] [airplane,\;person]\par\medskip

\textbf{\# Your Task}\par
Now, given the question below, list the entities that appear in the question.\par\medskip

\textbf{[Question]} \{Question\}\par
\textbf{[Detect]}\par

\end{tcolorbox}

\vspace{\baselineskip}
\noindent
\textbf{(2) Perspective Change (Sec.~\ref{subsec:perspective_change})} --- Setting a Reference Perspective

\begin{tcolorbox}[
  breakable,
  width=\linewidth,
  colback=blue!3!white,
  colframe=blue!60!black,
  boxrule=0.6pt,
  arc=1mm,
  left=1.2mm,
  right=1.2mm,
  top=0.8mm,
  bottom=0.8mm,
]

\ttfamily\small

Given a question about spatial reasoning, we want to extract the \emph{perspective} of the question. If the question is from the camera's perspective, return ++camera++.\par\medskip

\# \textbf{Example}\par
[Question] From the woman's perspective, is the tree on the left or right?\par
[Perspective] ++woman++\par\medskip

\# \textbf{Your Task}\par
Given the question below, please specify the \emph{perspective} from which the question is asked.\par
You must return in the format: [Perspective] ++object\_name++\par\medskip

\textbf{[Question]} \{Question\}\par
\textbf{[Options]} obj1, obj2, ..., camera\par

\textbf{[Perspective]}
\end{tcolorbox}

\vspace{\baselineskip}
\noindent
\textbf{(3) Egocentric Rephrasing (Sec.~\ref{subsec_app:egocentric_rephrasing})}

\begin{tcolorbox}[
  breakable,
  width=\linewidth,
  colback=blue!3!white,
  colframe=blue!60!black,
  boxrule=0.6pt,
  arc=1mm,
  left=1.2mm,
  right=1.2mm,
  top=0.8mm,
  bottom=0.8mm,
]

\ttfamily\small

From a sentence with a perspective description, we need to remove the perspective description.\par\medskip

\# \textbf{Example}\par
[Question] From the car's perspective, which is on the right side: the person or the tree?\par
[Output] Which is on the right side: the person or the tree?\par\medskip

\# \textbf{Your Task}\par
Given the question below, please remove the perspective description.\par\medskip
\textbf{[Question]} \{Question\} \\
\textbf{[Output]}
\end{tcolorbox}

\vspace{\baselineskip}
\noindent
\textbf{(4) Perspective Prompting (Sec.~\ref{subsec:perspective_prompting})} --- Visual Prompt.

\begin{tcolorbox}[
  breakable,
  width=\linewidth,
  colback=blue!3!white,
  colframe=blue!60!black,
  boxrule=0.6pt,
  arc=1mm,
  left=1.2mm,
  right=1.2mm,
  top=0.8mm,
  bottom=0.8mm,
]

\ttfamily\small

This is an image of a 3D scene.\par\medskip

- The viewer is facing towards the object that is \emph{closest to the center}.\par
- A \emph{larger} object is closer to the viewer compared to a \emph{smaller} object. \\

\# \textbf{Task}\par
Based on the image, please answer the following question.\par\medskip

\textbf{\{Question\}}\par\medskip

Please only return the answer.
\end{tcolorbox}

\vspace{\baselineskip}
\noindent
\textbf{(5) Perspective Prompting (Sec.~\ref{subsec:perspective_prompting})} --- Numerical Prompt.

\begin{tcolorbox}[
  breakable,
  width=\linewidth,
  colback=blue!3!white,
  colframe=blue!60!black,
  boxrule=0.6pt,
  arc=1mm,
  left=1.2mm,
  right=1.2mm,
  top=0.8mm,
  bottom=0.8mm,
]

\ttfamily\small

Imagine that you are at the \{src\_obj\}'s position and facing where it is facing.\par
We have the coordinates of different objects in \{src\_obj\}'s coordinate system.\par\medskip

\# \textbf{Coordinate System}\par
- The origin is at the \{src\_obj\}'s position.\par
- The \{src\_obj\}'s facing direction is [0, 0, 1], which is aligned with the z-axis.\par
- The x-axis is to the right, the y-axis is up, and the z-axis is forward.\par\medskip

\# \textbf{Object Coordinates}\par
[...]\par\medskip

\# \textbf{Task}\par
Given the above \{src\_obj\}'s coordinate system and the object coordinates, please answer the following question:\par\medskip

\textbf{[Question]} \{Question\}
\end{tcolorbox}

\end{document}